%% file: root.tex
\documentclass{article}
\usepackage[preprint]{corl_2025}
\input{Support/packages}
\input{Support/commands}

\begin{document}

\input{Sections/00-titleandabstract}
\input{Sections/01-introduction}

\input{Sections/03-preliminaries}

\input{Sections/04-method}
\input{Sections/05-experiments}

\input{Sections/06-ablations}

\input{Sections/02-relatedwork}

\input{Sections/07-conclusions}
\newpage
\input{Sections/08-acknowledgement}
\bibliography{root}
\newpage
\appendix
\input{supp}

\end{document}

%% file: Support/packages.tex
\usepackage{graphicx}
\usepackage{amsmath} 
\usepackage{amssymb}  
\usepackage{algorithm}
\usepackage{hyperref}
\usepackage[noend]{algpseudocode}
\usepackage[capitalize]{cleveref}
\usepackage{multirow}
\usepackage{booktabs}
\usepackage{multicol}
\usepackage{booktabs}
\usepackage{wrapfig}
\usepackage{subcaption}
\usepackage{caption}

%% file: Support/commands.tex
\crefname{section}{Sec.}{Secs.}
\Crefname{section}{Section}{Sections}
\Crefname{table}{Table}{Tables}
\crefname{table}{Tab.}{Tabs.}

%% file: Sections/00-titleandabstract.tex
\title{
    \LARGE \bf
    Real-is-Sim: Bridging the Sim-to-Real Gap with a Dynamic Digital Twin
}

\author{%
\begin{tabular}{c}
Jad Abou-Chakra$^{1}$, Lingfeng Sun$^{1}$, Krishan Rana$^{2}$, Brandon May$^{1}$,\\ Karl Schmeckpeper$^{1}$, Niko Suenderhauf$^{2}$, Maria Vittoria Minniti$^{1}$, Laura Herlant$^{1}$\\
\small{$^{1}$Robotics and AI Institute, $^{2}$Queensland University of Technology}
\end{tabular}%
}

\maketitle
\thispagestyle{empty}
\pagestyle{empty}

\begin{abstract}
We introduce real-is-sim, a new approach to integrating simulation into behavior cloning pipelines. In contrast to real-only methods, which lack the ability to safely test policies before deployment, and sim-to-real methods, which require complex adaptation to cross the sim-to-real gap, our framework allows policies to seamlessly switch between running on real hardware and running in parallelized virtual environments.
At the center of real-is-sim is a dynamic digital twin, powered by the Embodied Gaussian simulator, that synchronizes with the real world at 60Hz. This twin acts as a mediator between the behavior cloning policy and the real robot. Policies are trained using representations derived from \emph{simulator} states and always act on the \emph{simulated} robot, never the real one. During deployment, the real robot simply follows the simulated robot’s joint states,  and the simulation is continuously corrected with real world measurements. This setup, where the simulator drives all policy execution and maintains real-time synchronization with the physical world, shifts the responsibility of crossing the sim-to-real gap to the digital twin's synchronization mechanisms, instead of the policy itself. We demonstrate real-is-sim on a long-horizon manipulation task (PushT), showing that virtual evaluations are consistent with real-world results. We further show how real-world data can be augmented with virtual rollouts and compare to policies trained on different representations derived from the simulator state including object poses and rendered images from both static and robot-mounted cameras. Our results highlight the flexibility of the real-is-sim framework across training, evaluation, and deployment stages. Videos available at \href{https://real-is-sim.github.io}{https://real-is-sim.github.io}.

\end{abstract}

%% file: Sections/01-introduction.tex
\section{INTRODUCTION}

\noindent 
Simulation environments are a powerful tool for developing robotic manipulation policies. They offer key advantages such as continuous performance monitoring, large-scale parallel rollouts, full access to state information, and precise control over resets and environmental variation. These features make policy learning in simulation significantly more practical to do efficiently. However, translating these advantages to the real world is challenging. The main difficulty lies in ensuring that a policy behaves the same way in both simulation and reality. This requires two things: aligning the input distributions the policy sees in both domains, and ensuring that both the simulated and real robots respond similarly to the same policy outputs. Achieving both reliably is difficult.

\begin{figure}[t!]
  \centering
  \includegraphics[scale=1.0]{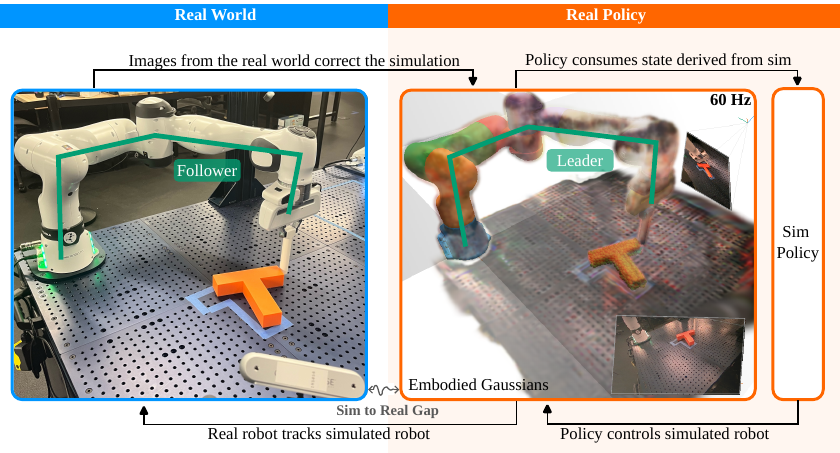}
  \caption{
  The real-is-sim framework, illustrating the information flow between its components. A policy trained in a physics simulator that can be synchronized with the real world~\cite{embodiedgaussians} controls a simulated robot. Real-world observations continuously update the simulator, maintaining its state close to ground truth. The real robot then mirrors the simulated robot's joint positions. This approach shifts the sim-to-real gap challenge from the policy to the physics simulator.
   }
  \label{fig-main}
  \vspace{-1em}
\end{figure}

To overcome this limitation, we introduce real-is-sim, a new paradigm in which a simulator that continuously synchronizes with the real world is the sole interface to a behavior cloning policy and is ever-present; even during real-world execution. This paradigm is powered by Embodied Gaussians~\cite{embodiedgaussians}, the first simulator to achieve real-time synchronization with physical environments in a generalizable way, through the integration of Gaussian splatting~\cite{gaussiansplating} and particle-based physics~\cite{xpbd}. This simulator enables the creation of a dynamic digital twin of the physical system that can be updated in real time via sensor feedback. Policies are trained on representations derived from the simulator's state, effectively reframing real-world interactions as simulated experiences and unifying training and deployment.

Real-is-sim functions by collecting synchronized simulation states during real-world interactions, which serve as demonstration trajectories for policy training. We avoid sim-only demonstrations to ensure all trajectories reflect motions the real robot can reliably execute whilst keeping the simulation synchronized with the real world. These policies operate on simulator-derived representations and produce joint-space actions for the simulated robot. This process is consistent in the two main operating modes: (i) when deployed on a real robot, and (ii) when evaluating in virtual environments. When connected to a real robot, the physical robot mirrors the simulated robot’s configuration, while sensor data continuously updates the state of the objects in the simulator. This architecture decouples the policy from the physical hardware, ensuring the policy always operates on simulator states drawn from a similar distribution as the training data. The real-is-sim paradigm enables seamless switching between real-world deployment and virtual evaluation, as the only distinction between the two modes is the activation or deactivation of the real robot’s follower behavior.

By maintaining a real-time digital twin, the real-is-sim framework unifies simulation and reality, enabling consistent policy execution, scalable offline evaluation, and flexible representation learning. Our experiments on the PushT task demonstrate three key applications of real-is-sim: (i) enabling rapid policy evaluation in a virtual environment that correlates strongly with real-world performance, (ii) supporting data augmentation by intermixing real and simulated data to improve performance, and (iii) identifying optimal simulation-derived representations to maximize task effectiveness.

%% file: Sections/03-preliminaries.tex
\section{PRELIMINARIES}

\subsection{Digital Twin}
The real-is-sim framework is built around a simulator that stays synchronized with the real world. This is achieved using \textit{Embodied Gaussians}~\cite{embodiedgaussians}, a simulator that continuously corrects itself using RGB-based feedback from real-world observations. The system represents the environment using two tightly coupled components: 
\begin{enumerate}
    \item \textbf{Oriented particles}, which capture physical properties such as position, rotation, mass, and radius. These are simulated using a particle-based physics system~\cite{xpbd}.
    \item \textbf{3D Gaussians}, which model visual appearance through parameters such as position, rotation, scale, color, and opacity. These are rendered using Gaussian splatting~\cite{gaussiansplating}.
\end{enumerate}
\begin{figure*}[t]
  \centering
  \includegraphics[width=\linewidth]{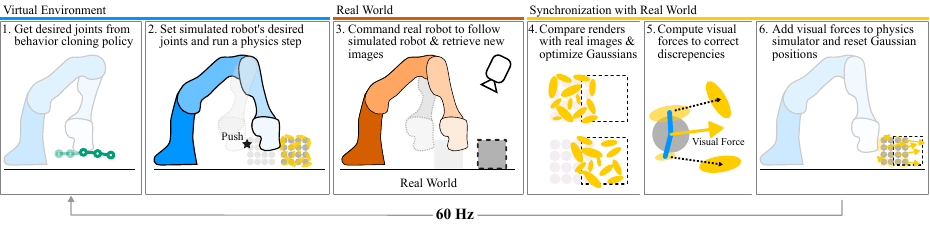}
  \caption{
Figure adapted from~\cite{embodiedgaussians}. It illustrates how Embodied Gaussians represents the world using physical particles and visual Gaussians. The Gaussians are corrected using real RGB images, and in turn exert fictitious visual forces on the particles to align the simulation with the real world. The adaptation highlights how this mechanism integrates into the real-is-sim framework.
  }
  \label{fig-embodied}
  \vspace{-1em}
\end{figure*}
Each particle can be rigidly linked to multiple Gaussians, allowing the physical and visual representations to move together. As the physics system updates the particles, the attached Gaussians follow accordingly. The key innovation in Embodied Gaussians is its correction mechanism based on photometric error. By comparing real camera images with rendered views, the system computes fictitious visual forces that act on the particles. These forces continuously adjust the simulation to stay aligned with reality. This correction loop runs at 60Hz, ensuring the simulator remains both physically consistent and visually accurate. An illustration of the corrective mechanism is provided in~\Cref{fig-embodied}. Additional details are available in~\cite{embodiedgaussians}.

\subsection{Behaviour Cloning}

In behavior cloning (BC), the goal is to learn a policy $\pi_\theta$ parameterized by $\theta$ that maps an observation $o_t$ at time step $t$ to a sequence of future actions $A_t = [a_t, a_{t+1}, \ldots, a_{t+H-1}]$ over a horizon $H$. Given a dataset of $N$ expert demonstrations $\mathcal{D} = \{(o^i, A^i)\}_{i=1}^N$, where $o^i = \{o_1^i, \ldots, o_T^i\}$ is a sequence of observations and $A^i = \{a_1^i, \ldots, a_T^i\}$ is the corresponding sequence of actions, the policy is trained to model the conditional distribution $p(A_t \mid o_t)$. This is typically done by minimizing the behavior cloning loss:
\begin{equation}
    \mathcal{L}_{\text{BC}}(\pi_\theta) = \mathbb{E}_{(o, A) \sim \mathcal{D}} \left[ \left\| \pi_\theta(o) - A \right\|^2 \right],
\end{equation}
where the policy’s predicted action sequence is penalized based on its deviation from the expert-provided action sequence.

Modeling temporally extended action sequences rather than single-step actions enables smoother and more consistent control~\cite{zhao2023learning,chi2023diffusion}. Our approach adopts \textit{Conditional Flow Matching} (CFM)~\cite{lipman2023flowmatchinggenerativemodeling}, a generative modeling method that learns a continuous trajectory distribution over actions. CFM models a velocity field $v^\theta(A_t^\tau \mid o_t)$ conditioned on $o_t$, predicting intermediate velocities along a path $A_t^\tau$ between a zero trajectory and the ground-truth $A_t$. The training objective minimizes the conditional flow-matching loss:
\begin{equation}
    \mathcal{L}_{\text{CFM}} = \mathbb{E}_{\tau,\, p(A_t \mid o_t),\, q_r(A_t^\tau \mid A_t)} \left\| u(A_t^\tau \mid A_t) - v^\theta(A_t^\tau \mid o_t) \right\|^2,
\end{equation}
where $\tau \sim \mathcal{U}(0,1)$ is a uniformly sampled interpolation timestep, $q_r(A_t^\tau \mid A_t)$ defines a reference trajectory (e.g., a Gaussian perturbation around a linear interpolation), and $u(\cdot)$ denotes the corresponding ground-truth velocity.

%% file: Sections/04-method.tex
\begin{figure*}[t]
  \centering
  \includegraphics[width=\linewidth]{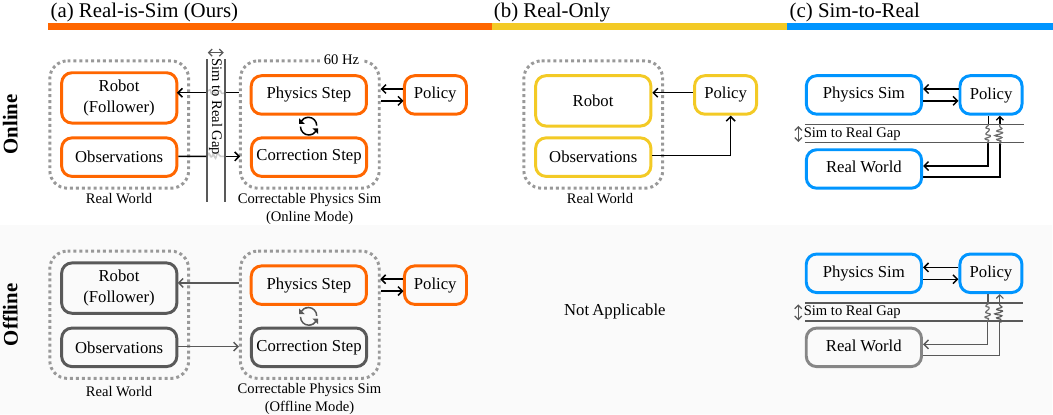}
  \caption{Comparison of three paradigms: (a) Real-is-Sim, (b) Real-Only, and (c) Sim-to-Real, highlighting their online (real-world interaction) and offline (simulation-based) capabilities.  Real-is-Sim offers a unified framework where deploying in a virtual environment is a simplified version of deploying in the real-world, lacking only real-time correction.  This ensures seamless transferability. Real-Only is confined to real-world execution.  Sim-to-Real struggles with distribution mismatch and dynamics discrepancies, making successful transfer uncertain.
  }
  \label{fig-frameworks}
  \vspace{-1em}
\end{figure*}

\section{METHOD}
Real-is-sim is a framework that uses a simulator as a mediator between the real world and a learnt policy. For this approach to work, the simulator must continuously synchronize with the physical world by updating its state based on real-world observations. In our work, we use Embodied Gaussians as the simulator, as it provides the necessary mechanisms for real-time alignment. This section first formalizes the framework, then details the complete pipeline: creating a scene, collecting demonstrations, training and deploying a policy, and evaluating the resulting policy.

\subsection{Framework}

\begin{wrapfigure}{r}{0.4\linewidth}
  \centering
  \includegraphics[width=\linewidth]{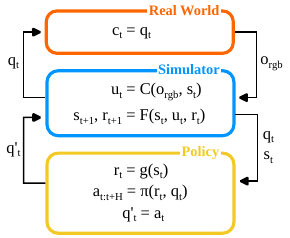}
  \caption{
  Information flow between subsystems.
  }
  \vspace{-1em}
  \label{fig-flow}
\end{wrapfigure}

Our framework consists of three subsystems: (i) the real-world system, (ii) the simulator, and (iii) the learned policy. Real-is-sim distinguishes between two modes: \textbf{online}, where the simulator is connected to and synchronized with the real-world system, and \textbf{offline}, where the system runs entirely in simulation and can be parallelized. We illustrate the connections in~\Cref{fig-frameworks}a and ~\Cref{fig-flow}. 

The \textbf{real-world system} is composed of a set of cameras that output RGB images, denoted as $o_\text{rgb}$, and a high-impedance joint controller that receives as input a desired joint configuration $c \in \mathbb{R}^d$ for a robot with $d$ degrees of freedom. 

The \textbf{simulator} maintains an environment state $s_t$, a simulated robot state $r_t$, an environment corrective input $u_t$, and a simulated robot control input $q'_t$ at each timestep $t$. These are defined as
\[
s_t = \{ p^i_t, R^i_t, v^i_t, \omega^i_t \}_{i=1}^{N_\text{obj}},
\quad
u_t = \{ f^i_t, \tau^i_t \}_{i=1}^{N_\text{obj}},
\quad
r_t = \{q_t, \dot{q}_t\}.
\]

\vspace{-0.5em}
For each simulated object $i \in {1, \dots, N_\text{obj}}$, the state is described by the position $p \in \mathbb{R}^3$, orientation $R \in SO(3)$, linear velocity $v \in \mathbb{R}^3$, and angular velocity $\omega \in \mathbb{R}^3$. The corrective control inputs, which align the simulated state with the real world, are computed as virtual forces $f \in \mathbb{R}^3$ and torques $\tau \in \mathbb{R}^3$ applied to each object.  For the simulated robot, the state includes the joint positions $q \in \mathbb{R}^d$, joint velocities $\dot{q} \in \mathbb{R}^d$, and the desired joint configuration $q' \in \mathbb{R}^d$. The next states $s_{t+1}$ and $r_{t+1}$ are computed through the simulator’s physics step: $s_{t+1}, r_{t+1} = F(s_t, u_t, r_t)$. If the real-world system is connected, the robot's desired joint configuration $c$ is set to $q_t$ at each timestep and the corrective inputs $u_t$ are set to $u_t = C(s_t, o_\text{rgb})$, where $C$ represents the corrective mechanism in Embodied Gaussians (illustrated in~\cref{fig-embodied}).

The \textbf{policy} is a function $\pi(r_t, q_t)$ that maps a representation $r_t = g(s_t)$ to a trajectory $a_{t:t+H} \in \mathbb{R}^{m \times (d+1)}$, where $r_t$ is derived from the simulator state $s_t$, $m$ is the number of waypoints in the trajectory (also called the action horizon). Each waypoint $a$ consists of a desired joint configuration $q' \in \mathbb{R}^d$ and the estimated progress towards task completion $p \in [0, 1]$. Once the trajectory is computed, the simulated robot follows it over a time horizon $H$, updating its desired joint configuration $q'_\text{desired}$ every $H/m$ seconds to reach the next waypoint in the trajectory $a$. After the trajectory is completed, a new one is generated. In our implementation, we set $H = 1~\text{second}$ and $m = 32$.

\subsection{Pipeline}
\textbf{Scene Setup} We construct the environment following the Embodied Gaussians procedure. Each object is generated from 4 viewpoints and is ultimately represented by a position $p$, a rotation $R$, and a set of fixed-radius spheres for collision geometry. Visual appearance is modeled by attaching Gaussians to each body, optimized using Gaussian splatting~\cite{gaussiansplating}. The ground and robot geometry and appearance are manually specified. 

\textbf{Demonstration Collection} Demonstrations are collected with Embodied Gaussians continuously correcting the simulated state using live camera observations. Collecting with the simulator in-the-loop allows users to monitor synchronization quality and adapt their demonstrations to maintain it. To improve synchronization, during demonstrations we minimize occlusions by keeping the robot body out of camera views, move quickly when not contacting objects, slow down during interactions, and maintain ample clearance around object edges. Throughout each demonstration, we record the full sequence of simulation states at 60Hz. Although the simulation models rich physical properties, only the object poses and robot joint configurations change over time; all other physical parameters are initialized once and remain constant. Similarly, visual states (the Gaussians) are only stored at the initial timestep. A full list of static parameters is provided in~\cite{macklin2022warp}.
\begin{figure*}[t]
  \centering
  \includegraphics[width=\linewidth]{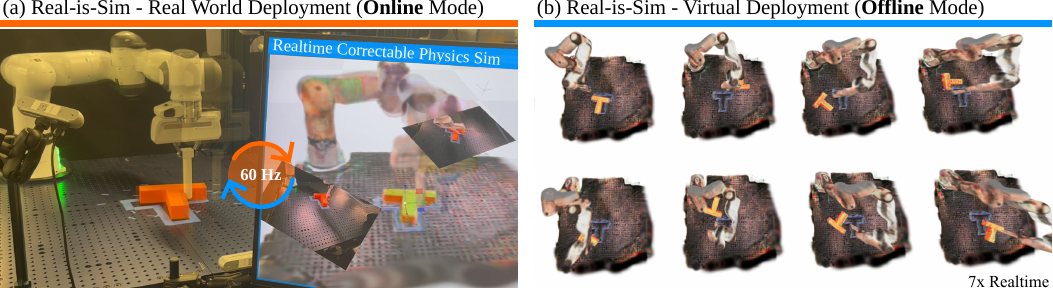}
  \caption{
The two modes of the Real-is-Sim framework: real-world deployment and virtual environment deployment.
  }
  \label{fig-onlineoffline}
  \vspace{-2em}
\end{figure*}

\begin{figure*}[t]

  \centering
  \includegraphics[width=\linewidth]{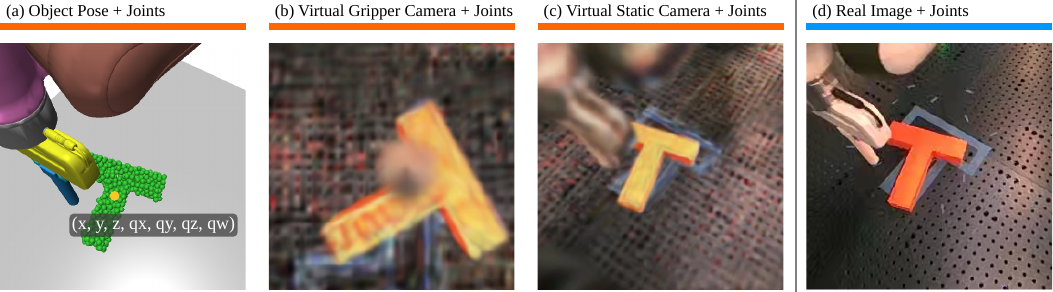}
  \caption{Comparison of representations from the Real-to-Sim framework versus a real image baseline: (a) State-based representation (object position and orientation), (b) Virtual camera on robot end effector, (c) Virtual camera at real camera position, (d) Real camera image, which is used to correct the simulation, but not as a direct input to the policy.
  }
  \label{fig-repr-visual}
  \vspace{-1em}
\end{figure*}

\textbf{Policy Training} Our policy outputs an action trajectory $a_{t:t+H}$, where each waypoint consists of a desired joint configuration $q' \in \mathbb{R}^d$ and a scalar progress value $p \in [0, 1]$. The progress value represents the relative completion of the task and is labeled based on the elapsed time in the demonstration, assuming that all demonstrations end at the target position. This heuristic, validated in~\cite{rana2024affordance}, allows us to evaluate policies without manually engineering rewards. The policy is conditioned on the robot’s current joint configuration $q_t$. For object state-based policies, we additionally condition on the object's pose, represented by a position and quaternion\footnote{The quaternion is used for simplicity and did not introduce ambiguity in our tasks.}. For image-based policies, we encode rendered images into 64-dimensional vectors using a ResNet. Embodied Gaussians enables the rendering of images from arbitrary viewpoints, allowing the creation of virtual cameras. The representations used for conditioning, including object pose and rendered images, are shown in~\Cref{fig-repr-visual}.

\textbf{Policy Deployment} During deployment, the simulator continues running alongside the real robot to ensure it remains synchronized with the physical world. The trained policy processes the state of the simulator and outputs the desired joint configurations for the simulated robot. This means the policy is operating with the same type of inputs it was trained on, controlling the simulated robot as it was during demonstration collection.
The integration with the real-world system occurs through two key mechanisms: (1) RGB images from the real-world cameras are continuously input into the system, providing real-time corrections to the state of the simulated objects. This feedback enables the simulator to continuously adjust and align the simulated environment with the real world. (2) The real robot's joint controller is configured to follow the simulated robot's joint positions $q$ (and not the desired joints $q'$). This design choice, where the real robot acts as a follower, allows for a smooth transition between operating on a real robot and operating solely in a virtual environment. 

\textbf{Policy Evaluation}  By deactivating the real-world coupling and corrective force calculation ($u_t$), the same setup can be used in a virtual-only environment. This allows for easy parallelization of environments which enables fast policy evaluation. In virtual deployments, we achieve a speedup of 5x to 8x, depending on the chosen representation for the policy, when running 20 environments concurrently.

%% file: Sections/05-experiments.tex
\section{EXPERIMENTS}
\begin{wrapfigure}{r}{0.5\linewidth}
  \centering
  \includegraphics[width=\linewidth]{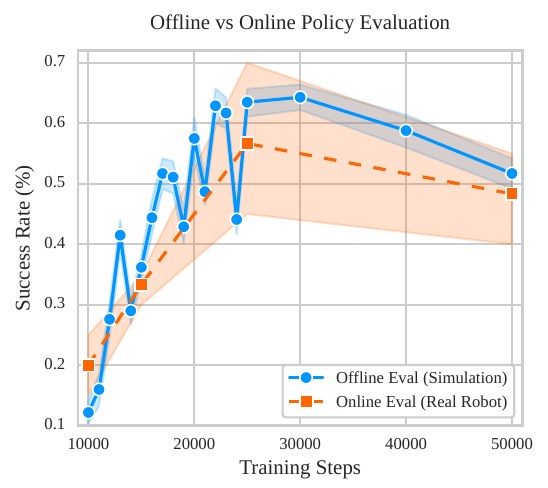}
  \caption{
Success rate of an imitation learning policy versus training steps, comparing evaluation on the real setup and in a virtual environment. The consistency between the two curves shows that the faster virtual-only evaluation is a reliable proxy for real-world testing, providing a practical alternative for policy assessment. Error bars represent 95\% confidence intervals.
  }
  \vspace{-1.5em}
  \label{fig-evalonlineoffline}
\end{wrapfigure}
\textbf{Experimental Setup} We evaluate our framework on the PushT task: a long horizon task where a robot uses a pusher to position a T-shaped object into a predefined location. The T-Block's geometry (represented as a set of spherical shapes attached to a single body frame) is automatically generated using the method from Embodied Gaussians \cite{embodiedgaussians}. We use three RealSense D455 cameras running at 90 fps with a resolution of $420 \times 270$. The robot mesh is known, and the ground plane is estimated from the camera-generated pointclouds. Gaussians attached to the robot, table, and T-Block are learned once using the Gaussian Splatting procedure outline in \cite{embodiedgaussians} and reused across experiments.

For object state-based policies, we use the pose of the object encoded as a translation vector $\mathbf{p} \in \mathbb{R}^3$ and a unit quaternion $\mathbf{R} \in \mathbb{SO}(3)$. For image-based policies, we use a 64 dimensional feature vector output by a ResNet~\cite{chi2023diffusion} which was jointly trained with the policy. The images input to the ResNet are rendered from the simulation using Gaussian Splatting from virtual cameras. These cameras can be placed anywhere. In our experiments, we use a static camera placed in the same location as the real camera and a virtual gripper camera which we mount to the simulated robot's end-effector (these are shown in~\Cref{fig-repr-visual}).

To evaluate policy success on the PushT task, we define 20 starting poses on the table that are not part of the demonstration set. In real-world testing, success is determined by whether the policy moves the T-Block to the target location from these poses. Each pose is tested three times, resulting in 60 measurements per success rate evaluation. In virtual-only testing, success is determined when the policy outputs a progress value above 0.9, a strategy verified in \cite{rana2024affordance}, which serves as a good proxy for success without requiring engineered reward values. 

\textbf{Offline Evaluation}
In this experiment, we demonstrate how real-is-sim's offline mode facilitates policy training and evaluation. A key challenge in imitation learning lies in selecting optimal checkpoints for deployment. This difficulty arises because the loss used in behavior cloning does not reliably indicate policy performance, and real-world robot evaluations require substantial time and resources. Consequently, researchers often default to using the final training checkpoints despite it being suboptimal.

We demonstrate that real-is-sim's offline mode enables efficient checkpoint selection through rapid evaluation. For a state-based policy trained on 30 demonstrations, we implement a flow-matching policy trained for 50,000 steps (\Cref{fig-evalonlineoffline}). We then compare two evaluation methodologies: (i) evaluation on only the virtual world (without the real robot attached and the corrective mechanisms engaged) and (ii) real-world online evaluations. Our virtual-only approach initializes environments with identical test cases to the real-world evaluations. The virtual-only evaluation calculates success rates through the policy-reported progress. While we use this metric to maintain generality, practitioners could implement more sophisticated reward functions for their specific tasks. 

We selected four representative checkpoints (10k, 15k, 25k, 50k) for real-world evaluation and performed more comprehensive analysis using faster offline evaluation. \Cref{fig-evalonlineoffline} shows that, while absolute performance may differ, the relative ordering of checkpoints is preserved - making offline evaluation a practical tool for selecting high-performing policies in minutes rather than hours.
Notably, later checkpoints do not consistently outperform earlier ones, suggesting practitioners could use offline evaluation for either early stopping or optimal checkpoint selection. While this experiment uses identical initial states for both evaluation modes to establish correlation, real applications could benefit from randomized initial states during training-phase evaluations.
\begin{figure}[t]
  \centering
  \includegraphics[width=\linewidth]{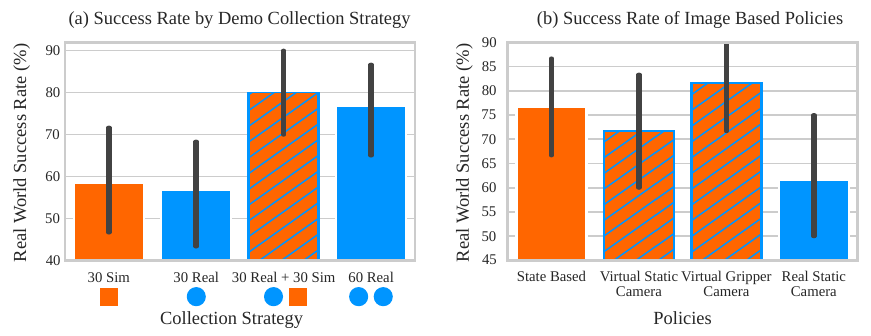}
  \caption{
(a) Success rates of different demonstration collection strategies. (b) Success rates of policies using different representations to condition the flow matching policy. Error bars are 95\% confidence intervals.
   }
   \label{fig-augmentation}
   \vspace{-1em}
\end{figure}

\textbf{Offline Data Collection and Augmentation}
In this experiment, we demonstrate how real-is-sim’s offline mode can be used to to collect additional demonstrations to increase task performance (\Cref{fig-augmentation}). We train a state-based policy using 30 real-world demonstrations collected in online mode - which achieves a success rate of around 57\%. We also augment the real-world demonstrations using the simulator only (offline mode).  To collect the additional demonstrations, we deploy this policy in offline mode across multiple parallel environments and identify failure states where the policy falls into a local minima. These failure states are then presented to the user to provide an additional 30 demonstrations specifically solving the task under these challenging configurations. By combining the original 30 real-world demonstrations with the 30 new simulated demonstrations, we retrain the policy and observe a significant improvement to 80\% success rate.

Lastly as a baseline, we also collect 30 demonstrations in simulation only. The resulting sim-only policy attains a similar success rate during deployment. The success rate of the offline only policy is comparable to the policy trained on real demonstrations. This is largely owing to the kinematic nature of the PushT task which means that successfully policy rollouts are mostly dependent on poses and not dynamic effects. This is precisely what Embodied Gaussians can easily correct upon deployment.  

This experiment highlights a straightforward yet powerful application of real-is-sim’s offline mode. The cost of augmenting the data used for policy training using real-is-sim's simulator is much lower than doing so in the real world. As a comparison, we also train a policy using 60 real-world demonstrations collected without augmentation. Interestingly, this approach yields a similar performance improvement as the augmented dataset, underscoring the effectiveness of real-is-sim’s offline mode in efficiently enhancing policy training.

\textbf{Representation Flexibility}
Another major benefit of real-is-sim’s offline mode is its flexibility in extracting diverse representations from the Embodied Gaussians simulation. For instance, it enables access to privileged information—such as precise object poses—that would be difficult to obtain in the real world. Additionally, virtual cameras can be freely positioned within the system, facilitating visuomotor policy learning from multiple perspectives. In our experiments, we demonstrate that policies leveraging different representations—including state-based object poses, a gripper-mounted virtual camera, and a static virtual camera—all successfully solve the task (\cref{fig-augmentation}b, \cref{fig-repr-visual}). Notably, these approaches achieve performance comparable to, and in some cases surpassing, the baseline policy that relies solely on real-world images and robot actions.

Interestingly, the policy using the gripper-mounted virtual camera achieves the highest performance (82\%) among all representations. This policy also exhibits behaviors such as as actively searching for the T-block when it leaves the camera’s field of view (see videos on our website). These findings align with prior work highlighting the advantages of wrist-mounted cameras over static ones in certain scenarios~\cite{hsu2022vision}. Within the real-to-sim framework, we can systematically evaluate different virtual camera configurations~\cite{goyal2023rvt} and their impact on policy learning—all using the same set of demonstrations. This flexibility allows us to identify the optimal representation for a given manipulation task in offline mode and seamlessly deploy it in online execution.

%% file: Sections/06-ablations.tex

%% file: Sections/02-relatedwork.tex
\section{RELATED WORK}
\textbf{Limitations of Real-World Policy Evaluation}: Evaluating robotic policies directly in the real world, while reliable, suffers from being labor-intensive, time-consuming \cite{zhou2025autoeval, kress2024robot}, and facing scalability bottlenecks due to manual processes \cite{chi2024diffusionpolicy}. Consequently, researchers explore simulation for safer, repeatable, and scalable evaluation \cite{kadian2020sim2real, pmlr-v270-li25c, silwal2024we, khanna2024habitat}, often focusing on improving simulation fidelity \cite{kadian2020sim2real, khanna2024habitat, silwal2024we, pmlr-v270-li25c}. Our framework differs from this approach by integrating a continuously synchronized simulator throughout the entire policy lifecycle (training, evaluation, deployment). This real-time alignment via sensor data enables direct deployment of simulation-trained policies without further tuning and ensures a unified observation and control space for scalable evaluation.

\textbf{Addressing the Sim-to-Real Gap in Robotic Manipulation}: Simulation offers a scalable alternative to real-world evaluation \cite{zhou2025autoeval}, but the sim-to-real gap remains a key challenge. Existing strategies include domain randomization to enhance robustness \cite{Tobin2017DomainRandom} and domain adaptation to align representations \cite{James2018SimtoRealVS}, as well as methods focusing on aligning simulation dynamics \cite{pmlr-v229-ren23b, pfaff2025scalable} or using real-to-sim-to-real approaches \cite{Villasevil-RSS-24, pmlr-v270-jiang25a}. Frameworks like DeepMind’s robot soccer also combine neural rendering with domain randomization \cite{tirumala2024soccer}.
While these methods aim to bridge the sim-to-real gap by making policies more robust or adapting representations, our "real-is-sim" paradigm eliminates this gap entirely. By continuously aligning the simulator with the real world and treating it as the execution environment, we decouple the policy from sim-real discrepancies. The challenge of domain alignment shifts to the simulator itself, which is addressed through real-time updates.

\textbf{Leveraging World Models for Robotics}: Learned world models enable robots to predict future outcomes for planning and control. These range from 2D predictors \cite{yang2024unisim} to more structured 3D physically grounded models, such as those utilizing Gaussian Splatting \cite{phystwin, embodiedgaussians}, which allow for real-time tracking and simulated interaction. Our work builds upon this by exploring the integration of such advanced 3D world models within a behavior cloning framework. This aims to harness the advantages of simulation directly in real-world tasks by training policies on the dynamically updated, real-time digital twin provided by these models.

%% file: Sections/07-conclusions.tex
\section{CONCLUSIONS}
We introduced real-is-sim, a unified framework for policy training, evaluation, and deployment that maintains a continuously corrected simulation throughout the entire pipeline. By treating simulation as a persistent, real-time interface, enabled by Embodied Gaussians, we eliminate the disconnect between training and deployment domains. This reframes policy evaluation as a simulation-native process, even in real-world settings. Our results on the PushT task show strong alignment between offline and real-world performance, demonstrating that real-is-sim enables scalable, efficient evaluation without sacrificing fidelity. We hope this framework serves as a foundation for future real-world learning algorithms that leverage this tight coupling between simulation and reality.

%% file: Sections/08-acknowledgement.tex
\section{Limitations}

While our framework demonstrates promising results, its applicability is fundamentally constrained by the fidelity of the underlying physical simulator. Scenarios that diverge significantly from the simulator’s modeling assumptions present challenges for both forward prediction and corrective mechanisms. Future work could address these limitations by improving simulator accuracy through adaptive parameter estimation—for example, real-time optimization of physical properties such as friction or mass, as explored in~\cite{phystwin} and~\cite{longhini2024clothsplatting}.

Another limitation of the current framework is its reliance on visual forces to correct the simulator state. This restricts applicability to scenarios where objects are frequently visible and do not undergo prolonged occlusion or manipulation without observation.

Moreover, for Embodied Gaussians to function effectively, visual corrections must not conflict significantly with the dynamics predicted by the physics simulator. The approach assumes that both the simulator and the visual feedback are aligned, that is, they attempt to move the object in compatible directions. When this assumption breaks down and the corrections diverge, synchronization between the simulation and the real world may not be reliable. However, this limitation can be addressed. The synchronization mechanism can be extended with additional corrective signals beyond visual forces. For example, semantic cues, such as correspondence detection via methods like TAPIR~\cite{tapir} or loss augmentation using segmentation masks as in~\cite{dynamicgaussians}, could provide more robust alignment in challenging settings. The current framework was designed to demonstrate the simplest viable instantiation. Future extensions could incorporate these more sophisticated correction signals to broaden applicability across diverse and visually ambiguous environments.

Another limitation lies in the scene initialization process. In this work, we employed a simple scheme that is sensitive to factors such as the number and placement of cameras, object complexity, and visibility. However, recent advances in learned priors for geometry initialization, such as InstantMesh~\cite{instantmesh} and InstantSplat~\cite{instantsplat}, offer promising alternatives that could replace or enhance our approach, expanding the set of objects that can be initialized effectively.

Although we focused on the PushT task to validate the core contributions, the proposed methodology is broadly applicable to other manipulation domains. Extending this approach to tasks such as multi-object rearrangement, cloth manipulation, or non-prehensile pushing will require adapting the simulator’s dynamics and corrective policies to accommodate more complex interactions.

\section*{ACKNOWLEDGMENT}
The QUT authors acknowledge continued support
from the Queensland University of Technology (QUT) through the Centre
for Robotics. Their work was partially supported by the Australian Government through the Australian Research Council’s Discovery Projects funding
scheme (Project DP220102398). 

%% file: supp.tex
\captionsetup[figure]{name=Supplementary Figure}
\setcounter{figure}{0} 
\captionsetup[table]{name=Supplementary Table}
\setcounter{table}{0} 
\crefname{figure}{Supp. Figure}{Supp. Figures}
\crefname{table}{Supp. Table}{Supp. Tables}

\begin{figure}[b]
  \centering
  \includegraphics[width=\linewidth]{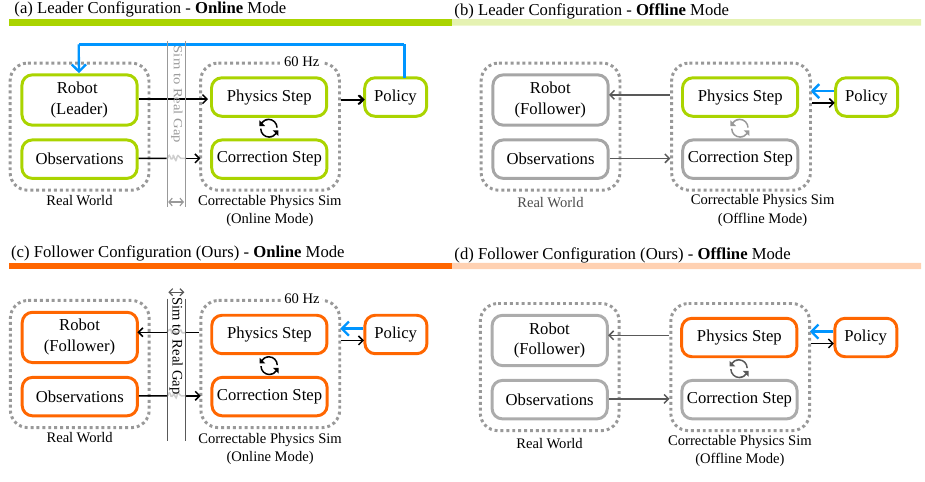}
  \caption{
Comparison of the real-is-sim framework in the proposed follower configuration (real robot mimics simulation) and an alternative leader configuration (simulation follows real robot). The follower setup requires fewer sim-to-real data streams and maintains identical system connections in both offline and online modes, simplifying deployment.
  }
  \label{fig-leader}
\end{figure}

\begin{center}
    \LARGE \textbf{
    Supplementary Material \\
    Real-is-Sim: Bridging the Sim-to-Real Gap with a Dynamic Digital Twin
    }
    
\end{center}

\section{Follower Configuration vs Leader Configuration}

We present real-is-sim as a framework that employs a continuously corrected digital twin of the environment to decouple a behavior cloning policy from the real robot. A central design decision in this framework is to configure the real robot as a \textit{follower} of the simulated robot. In this setup, when the real robot is connected, it directly mirrors the joint states of the simulated counterpart.

This follower design introduces a distinctive characteristic: the real world consistently lags slightly behind the simulated twin. To our knowledge, this temporal asymmetry is unique to \textit{Real-is-Sim} and implies that the system always possesses a short-horizon preview of future states. While we find this property conceptually interesting, our primary motivation for adopting the follower configuration is practical, it results in a significantly simpler software architecture and reduces the complexity of sim-to-real integration.

Supplementary \Cref{fig-leader} illustrates this distinction. In both offline and online modes under the follower configuration (our proposed setup), the system interconnections (blue lines) remain constant. In contrast, the \textit{leader} configuration—where the real robot dictates state and the simulator reacts—requires dynamic changes to the system architecture. Specifically, three data streams must cross the sim-to-real boundary in the leader case:
\begin{enumerate}
    \item Control inputs from the policy to the real robot,
    \item Control inputs from the real robot to the simulator,
    \item Observations from the real robot to the simulator.
\end{enumerate}

By comparison, the follower configuration requires only two such streams. This reduction in cross-boundary communication simplifies the overall system and motivates our recommendation of the follower configuration in \textit{Real-is-Sim}.

\section{Benefits of Collecting Demonstrations with the Real Robot}

\begin{wrapfigure}{r}{0.6\linewidth}
  \centering
  \includegraphics[width=\linewidth]{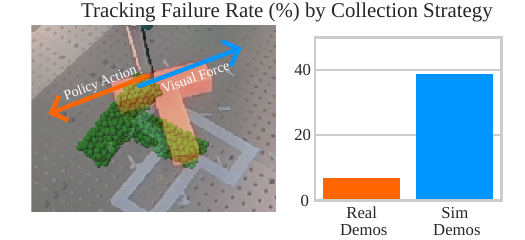}
  \caption{
Tracking failure in two policies, one trained on demonstrations collected with the real robot in the loop, and the other trained entirely in simulation. The left image illustrates a failure case where the simulated physics conflicts with the visual correction required to stay in sync with the real world. Since physics acts as a hard constraint, the correction cannot be applied. Collecting demonstrations in online mode helps avoid such failure modes by encouraging behaviors that maintain alignment between the simulation and reality.
  }
  \label{fig-tracking}
\end{wrapfigure}

In real-is-sim, we advocate collecting demonstrations with the real robot attached (online mode). During data collection, the user observes the state of the simulator overlaid with the real world, allowing them to continuously assess synchronization quality. This enables the user to adapt their behavior not only to complete the task but also to maintain alignment between the simulation and the real environment. Practically, this means knowing when to move quickly or slowly, and how much clearance to give during fast motions to accommodate slight mismatches between the two domains.

Supplementary \Cref{fig-tracking} illustrates an exaggerated case in which demonstrations are collected entirely in simulation and the resulting policy is deployed on the real robot. Although the task is successfully demonstrated in simulation, it is done with insufficient clearance around objects—causing impacts in the virtual world that do not occur in the real one. The perception system's correction mechanisms are unable to recover from this divergence, leading to a marked increase in tracking errors during real-world execution.

For this reason, we recommend that at least some demonstrations be collected with the real robot in the loop. This allows both the policy and the demonstrator to implicitly learn behaviors that help keep the simulator and real world in sync.

\section{Implementation Details}
We use an NVIDIA RTX 6000 Ada as our primary compute device. The robot employed is a Franka Emika with 7 degrees of freedom. The high-impedance controller for the real robot utilizes Ruckig~\cite{ruckig} to generate joint trajectories with minimal jerk, in conjunction with the Franka Emika's built-in joint controller. During demonstration collection, the simulated robot is constrained to act within a 2D plane using Pink~\cite{pink}. Note that this 2D Cartesian controller is only active during demo collection and only controls the simulated robot. The real robot continues to execute a joint impedance controller based on the simulated robot's joint positions. Interestingly, the policy learns to operate in 2D, even though the joint controller does not impose such a restriction when the policy is deployed.

\section{Alterations to Embodied Gaussians}
Our adaptation of Embodied Gaussians~\cite{embodiedgaussians} (built using~\cite{macklin2022warp}) differs from the original framework in two key aspects. First, we omit shape-matching constraints for deformable objects, modeling all entities as rigid bodies composed of spherical collision primitives. This results in a rigid body simulator rather than a particle-based one, improving performance from 30Hz to 60Hz in online mode. Second, we treat the robot as a dynamic entity, simulating mass and inertia for its links and joints, in contrast to the original kinematic formulation, where robot motion was purely kinematic. This change enhances simulation stability and allows the simulated robot to react to the environment. We also found it essential to ensure that the bodies comprising the simulated robot are not affected by gravity. Without this, controlling the simulated robot becomes more difficult, as the controller would need to include a gravity compensation term.

\section{Future Work}
While the always-in-the-loop simulator already corrects itself using observations, its accuracy could be further enhanced by backpropagating into physics parameters or learning a residual physics network, reducing reliance on explicit corrections by making the simulator’s predictions more physically faithful. A more accurate simulator would also enable reinforcement-learning-based fine-tuning of imitation-learned policies. Additionally, this work focused on offline policy evaluation, but future extensions could incorporate online planning during deployment for real-time adaptability. Finally, Embodied Gaussians could be extended to encode higher-level features such as DINO embeddings~\cite{oquab2023dinov2} or other semantic descriptors. This would enrich policy training and further make use of the simulator’s role as a mediator between the real world and the policy.

%% file: root.bbl
\begin{thebibliography}{34}
\providecommand{\natexlab}[1]{#1}
\providecommand{\url}[1]{\texttt{#1}}
\expandafter\ifx\csname urlstyle\endcsname\relax
  \providecommand{\doi}[1]{doi: #1}\else
  \providecommand{\doi}{doi: \begingroup \urlstyle{rm}\Url}\fi

\bibitem[Abou-Chakra et~al.(2024)Abou-Chakra, Rana, Dayoub, and
  Suenderhauf]{embodiedgaussians}
J.~Abou-Chakra, K.~Rana, F.~Dayoub, and N.~Suenderhauf.
\newblock Physically embodied gaussian splatting: A visually learnt and
  physically grounded 3d representation for robotics.
\newblock In \emph{8th Annual Conference on Robot Learning}, 2024.
\newblock URL \url{https://openreview.net/forum?id=AEq0onGrN2}.

\bibitem[Kerbl et~al.(2023)Kerbl, Kopanas, Leimk{\"u}hler, and
  Drettakis]{gaussiansplating}
B.~Kerbl, G.~Kopanas, T.~Leimk{\"u}hler, and G.~Drettakis.
\newblock 3d gaussian splatting for real-time radiance field rendering.
\newblock \emph{ACM Transactions on Graphics}, 42\penalty0 (4), July 2023.
\newblock URL \url{https://repo-sam.inria.fr/fungraph/3d-gaussian-splatting/}.

\bibitem[Macklin et~al.(2016)Macklin, M\"{u}ller, and Chentanez]{xpbd}
M.~Macklin, M.~M\"{u}ller, and N.~Chentanez.
\newblock Xpbd: Position-based simulation of compliant constrained dynamics.
\newblock In \emph{Proceedings of the 9th International Conference on Motion in
  Games}, MIG '16, page 49–54, New York, NY, USA, 2016. Association for
  Computing Machinery.
\newblock ISBN 9781450345927.
\newblock \doi{10.1145/2994258.2994272}.
\newblock URL \url{https://doi.org/10.1145/2994258.2994272}.

\bibitem[Zhao et~al.(2023)Zhao, Kumar, Levine, and Finn]{zhao2023learning}
T.~Zhao, V.~Kumar, S.~Levine, and C.~Finn.
\newblock Learning fine-grained bimanual manipulation with low-cost hardware.
\newblock \emph{Robotics: Science and Systems XIX}, 2023.

\bibitem[Chi et~al.(2023)Chi, Xu, Feng, Cousineau, Du, Burchfiel, Tedrake, and
  Song]{chi2023diffusion}
C.~Chi, Z.~Xu, S.~Feng, E.~Cousineau, Y.~Du, B.~Burchfiel, R.~Tedrake, and
  S.~Song.
\newblock Diffusion policy: Visuomotor policy learning via action diffusion.
\newblock \emph{The International Journal of Robotics Research}, page
  02783649241273668, 2023.

\bibitem[Lipman et~al.(2023)Lipman, Chen, Ben-Hamu, Nickel, and
  Le]{lipman2023flowmatchinggenerativemodeling}
Y.~Lipman, R.~T.~Q. Chen, H.~Ben-Hamu, M.~Nickel, and M.~Le.
\newblock Flow matching for generative modeling, 2023.
\newblock URL \url{https://arxiv.org/abs/2210.02747}.

\bibitem[Macklin(2022)]{macklin2022warp}
M.~Macklin.
\newblock Warp: A high-performance python framework for gpu simulation and
  graphics.
\newblock In \emph{NVIDIA GPU Technology Conference (GTC)}, 2022.

\bibitem[Rana et~al.(2024)Rana, Abou-Chakra, Garg, Lee, Reid, and
  Suenderhauf]{rana2024affordance}
K.~Rana, J.~Abou-Chakra, S.~Garg, R.~Lee, I.~Reid, and N.~Suenderhauf.
\newblock Affordance-centric policy learning: Sample efficient and
  generalisable robot policy learning using affordance-centric task frames.
\newblock \emph{arXiv preprint arXiv:2410.12124}, 2024.

\bibitem[Hsu et~al.(2022)Hsu, Kim, Rafailov, Wu, and Finn]{hsu2022vision}
K.~Hsu, M.~J. Kim, R.~Rafailov, J.~Wu, and C.~Finn.
\newblock Vision-based manipulators need to also see from their hands.
\newblock In \emph{ICLR}, 2022.

\bibitem[Goyal et~al.(2023)Goyal, Xu, Guo, Blukis, Chao, and Fox]{goyal2023rvt}
A.~Goyal, J.~Xu, Y.~Guo, V.~Blukis, Y.-W. Chao, and D.~Fox.
\newblock Rvt: Robotic view transformer for 3d object manipulation.
\newblock In \emph{Conference on Robot Learning}, pages 694--710. PMLR, 2023.

\bibitem[Zhou et~al.(2025)Zhou, Atreya, Tan, Pertsch, and
  Levine]{zhou2025autoeval}
Z.~Zhou, P.~Atreya, Y.~L. Tan, K.~Pertsch, and S.~Levine.
\newblock Autoeval: Autonomous evaluation of generalist robot manipulation
  policies in the real world.
\newblock \emph{arXiv preprint arXiv:2503.24278}, 2025.

\bibitem[Kress-Gazit et~al.(2024)Kress-Gazit, Hashimoto, Kuppuswamy, Shah,
  Horgan, Richardson, Feng, and Burchfiel]{kress2024robot}
H.~Kress-Gazit, K.~Hashimoto, N.~Kuppuswamy, P.~Shah, P.~Horgan, G.~Richardson,
  S.~Feng, and B.~Burchfiel.
\newblock Robot learning as an empirical science: Best practices for policy
  evaluation.
\newblock \emph{arXiv preprint arXiv:2409.09491}, 2024.

\bibitem[Chi et~al.(2024)Chi, Xu, Feng, Cousineau, Du, Burchfiel, Tedrake, and
  Song]{chi2024diffusionpolicy}
C.~Chi, Z.~Xu, S.~Feng, E.~Cousineau, Y.~Du, B.~Burchfiel, R.~Tedrake, and
  S.~Song.
\newblock Diffusion policy: Visuomotor policy learning via action diffusion.
\newblock \emph{The International Journal of Robotics Research}, 2024.

\bibitem[Kadian et~al.(2020)Kadian, Truong, Gokaslan, Clegg, Wijmans, Lee,
  Savva, Chernova, and Batra]{kadian2020sim2real}
A.~Kadian, J.~Truong, A.~Gokaslan, A.~Clegg, E.~Wijmans, S.~Lee, M.~Savva,
  S.~Chernova, and D.~Batra.
\newblock Sim2real predictivity: Does evaluation in simulation predict
  real-world performance?
\newblock \emph{IEEE Robotics and Automation Letters}, 5\penalty0 (4):\penalty0
  6670--6677, 2020.

\bibitem[Li et~al.(2025)Li, Hsu, Gu, Mees, Pertsch, Walke, Fu, Lunawat, Sieh,
  Kirmani, Levine, Wu, Finn, Su, Vuong, and Xiao]{pmlr-v270-li25c}
X.~Li, K.~Hsu, J.~Gu, O.~Mees, K.~Pertsch, H.~R. Walke, C.~Fu, I.~Lunawat,
  I.~Sieh, S.~Kirmani, S.~Levine, J.~Wu, C.~Finn, H.~Su, Q.~Vuong, and T.~Xiao.
\newblock Evaluating real-world robot manipulation policies in simulation.
\newblock In P.~Agrawal, O.~Kroemer, and W.~Burgard, editors, \emph{Proceedings
  of The 8th Conference on Robot Learning}, volume 270 of \emph{Proceedings of
  Machine Learning Research}, pages 3705--3728. PMLR, 06--09 Nov 2025.

\bibitem[Silwal et~al.(2024)Silwal, Yadav, Wu, Vakil, Majumdar, Arnaud, Chen,
  Berges, Batra, Rajeswaran, et~al.]{silwal2024we}
S.~Silwal, K.~Yadav, T.~Wu, J.~Vakil, A.~Majumdar, S.~Arnaud, C.~Chen, V.-P.
  Berges, D.~Batra, A.~Rajeswaran, et~al.
\newblock What do we learn from a large-scale study of pre-trained visual
  representations in sim and real environments?
\newblock In \emph{2024 IEEE International Conference on Robotics and
  Automation (ICRA)}, pages 17515--17521. IEEE, 2024.

\bibitem[Khanna et~al.(2024)Khanna, Mao, Jiang, Haresh, Shacklett, Batra,
  Clegg, Undersander, Chang, and Savva]{khanna2024habitat}
M.~Khanna, Y.~Mao, H.~Jiang, S.~Haresh, B.~Shacklett, D.~Batra, A.~Clegg,
  E.~Undersander, A.~X. Chang, and M.~Savva.
\newblock Habitat synthetic scenes dataset (hssd-200): An analysis of 3d scene
  scale and realism tradeoffs for objectgoal navigation.
\newblock In \emph{Proceedings of the IEEE/CVF Conference on Computer Vision
  and Pattern Recognition}, pages 16384--16393, 2024.

\bibitem[Tobin et~al.(2017)Tobin, Fong, Ray, Schneider, Zaremba, and
  Abbeel]{Tobin2017DomainRandom}
J.~Tobin, R.~Fong, A.~Ray, J.~Schneider, W.~Zaremba, and P.~Abbeel.
\newblock Domain randomization for transferring deep neural networks from
  simulation to the real world.
\newblock In \emph{IEEE/RSJ International Conference on Intelligent Robots and
  Systems (IROS)}, 2017.

\bibitem[James et~al.(2019)James, Wohlhart, Kalakrishnan, Kalashnikov, Irpan,
  Ibarz, Levine, Hadsell, and Bousmalis]{James2018SimtoRealVS}
S.~James, P.~Wohlhart, M.~Kalakrishnan, D.~Kalashnikov, A.~Irpan, J.~Ibarz,
  S.~Levine, R.~Hadsell, and K.~Bousmalis.
\newblock Sim-to-real via sim-to-sim: Data-efficient robotic grasping via
  randomized-to-canonical adaptation networks.
\newblock In \emph{IEEE/CVF Conference on Computer Vision and Pattern
  Recognition (CVPR)}, 2019.

\bibitem[Ren et~al.(2023)Ren, Dai, Burchfiel, and Majumdar]{pmlr-v229-ren23b}
A.~Z. Ren, H.~Dai, B.~Burchfiel, and A.~Majumdar.
\newblock Adaptsim: Task-driven simulation adaptation for sim-to-real transfer.
\newblock In J.~Tan, M.~Toussaint, and K.~Darvish, editors, \emph{Proceedings
  of The 7th Conference on Robot Learning}, volume 229 of \emph{Proceedings of
  Machine Learning Research}, pages 3434--3452. PMLR, 06--09 Nov 2023.

\bibitem[Pfaff et~al.(2025)Pfaff, Fu, Binagia, Isola, and
  Tedrake]{pfaff2025scalable}
N.~Pfaff, E.~Fu, J.~Binagia, P.~Isola, and R.~Tedrake.
\newblock Scalable real2sim: Physics-aware asset generation via robotic
  pick-and-place setups.
\newblock arXiv preprint arXiv:2503.00370, 2025.

\bibitem[Villasevil et~al.(2024)Villasevil, Simeonov, Li, Chan, Chen, Gupta,
  and Agrawal]{Villasevil-RSS-24}
M.~T. Villasevil, A.~Simeonov, Z.~Li, A.~Chan, T.~Chen, A.~Gupta, and
  P.~Agrawal.
\newblock Reconciling reality through simulation: A real-to-sim-to-real
  approach for robust manipulation.
\newblock In \emph{Proceedings of Robotics: Science and Systems}, Delft,
  Netherlands, July 2024.
\newblock \doi{10.15607/RSS.2024.XX.015}.

\bibitem[Jiang et~al.(2025)Jiang, Wang, Zhang, Wu, and
  Fei-Fei]{pmlr-v270-jiang25a}
Y.~Jiang, C.~Wang, R.~Zhang, J.~Wu, and L.~Fei-Fei.
\newblock Transic: Sim-to-real policy transfer by learning from online
  correction.
\newblock In P.~Agrawal, O.~Kroemer, and W.~Burgard, editors, \emph{Proceedings
  of The 8th Conference on Robot Learning}, volume 270 of \emph{Proceedings of
  Machine Learning Research}, pages 1691--1729. PMLR, 06--09 Nov 2025.

\bibitem[Tirumala et~al.(2024)Tirumala, Wulfmeier, Moran, Huang, Humplik,
  Lever, Haarnoja, Hasenclever, Byravan, Batchelor, Sreendra, Patel, Gwira,
  Nori, Riedmiller, and Heess]{tirumala2024soccer}
D.~Tirumala, M.~Wulfmeier, B.~Moran, S.~Huang, J.~Humplik, G.~Lever,
  T.~Haarnoja, L.~Hasenclever, A.~Byravan, N.~Batchelor, N.~Sreendra, K.~Patel,
  M.~Gwira, F.~Nori, M.~Riedmiller, and N.~Heess.
\newblock Learning robot soccer from egocentric vision with deep reinforcement
  learning.
\newblock arXiv preprint arXiv:2405.02425, 2024.

\bibitem[Yang et~al.(2024)Yang, Du, Ghasemipour, Tompson, Kaelbling,
  Schuurmans, and Abbeel]{yang2024unisim}
S.~S. Yang, Y.~Du, K.~Ghasemipour, J.~Tompson, L.~Kaelbling, D.~Schuurmans, and
  P.~Abbeel.
\newblock Learning interactive real-world simulators.
\newblock arXiv preprint arXiv:2310.06114, 2024.

\bibitem[Jiang et~al.(2025)Jiang, Hsu, Zhang, Yu, Wang, and Li]{phystwin}
H.~Jiang, H.-Y. Hsu, K.~Zhang, H.-N. Yu, S.~Wang, and Y.~Li.
\newblock Phystwin: Physics-informed reconstruction and simulation of
  deformable objects from videos.
\newblock \emph{arXiv preprint arXiv:2503.17973}, 2025.

\bibitem[Longhini et~al.(2024)Longhini, B{\"u}sching, Duisterhof, Lundell,
  Ichnowski, Bj{\"o}rkman, and Kragic]{longhini2024clothsplatting}
A.~Longhini, M.~B{\"u}sching, B.~P. Duisterhof, J.~Lundell, J.~Ichnowski,
  M.~Bj{\"o}rkman, and D.~Kragic.
\newblock Cloth-splatting: 3d cloth state estimation from {RGB} supervision.
\newblock In \emph{8th Annual Conference on Robot Learning}, 2024.
\newblock URL \url{https://openreview.net/forum?id=WmWbswjTsi}.

\bibitem[Doersch et~al.(2023)Doersch, Yang, Vecerik, Gokay, Gupta, Aytar,
  Carreira, and Zisserman]{tapir}
C.~Doersch, Y.~Yang, M.~Vecerik, D.~Gokay, A.~Gupta, Y.~Aytar, J.~Carreira, and
  A.~Zisserman.
\newblock {TAPIR}: Tracking any point with per-frame initialization and
  temporal refinement.
\newblock In \emph{Proceedings of the IEEE/CVF International Conference on
  Computer Vision}, pages 10061--10072, 2023.

\bibitem[Luiten et~al.(2024)Luiten, Kopanas, Leibe, and
  Ramanan]{dynamicgaussians}
J.~Luiten, G.~Kopanas, B.~Leibe, and D.~Ramanan.
\newblock Dynamic 3d gaussians: Tracking by persistent dynamic view synthesis.
\newblock In \emph{3DV}, 2024.

\bibitem[Xu et~al.(2024)Xu, Cheng, Gao, Wang, Gao, and Shan]{instantmesh}
J.~Xu, W.~Cheng, Y.~Gao, X.~Wang, S.~Gao, and Y.~Shan.
\newblock Instantmesh: Efficient 3d mesh generation from a single image with
  sparse-view large reconstruction models.
\newblock \emph{arXiv preprint arXiv:2404.07191}, 2024.

\bibitem[Fan et~al.(2024)Fan, Wen, Cong, Wang, Zhang, Ding, Xu, Ivanovic,
  Pavone, Pavlakos, Wang, and Wang]{instantsplat}
Z.~Fan, K.~Wen, W.~Cong, K.~Wang, J.~Zhang, X.~Ding, D.~Xu, B.~Ivanovic,
  M.~Pavone, G.~Pavlakos, Z.~Wang, and Y.~Wang.
\newblock Instantsplat: Sparse-view gaussian splatting in seconds, 2024.

\bibitem[Berscheid and Kr{\"o}ger(2021)]{ruckig}
L.~Berscheid and T.~Kr{\"o}ger.
\newblock Jerk-limited real-time trajectory generation with arbitrary target
  states.
\newblock \emph{Robotics: Science and Systems XVII}, 2021.

\bibitem[Caron et~al.(2025)Caron, De~Mont-Marin, Budhiraja, Bang, Domrachev,
  and Nedelchev]{pink}
S.~Caron, Y.~De~Mont-Marin, R.~Budhiraja, S.~H. Bang, I.~Domrachev, and
  S.~Nedelchev.
\newblock {Pink: Python inverse kinematics based on Pinocchio}, 2025.
\newblock URL \url{https://github.com/stephane-caron/pink}.

\bibitem[Oquab et~al.(2023)Oquab, Darcet, Moutakanni, Vo, Szafraniec, Khalidov,
  Fernandez, Haziza, Massa, El-Nouby, et~al.]{oquab2023dinov2}
M.~Oquab, T.~Darcet, T.~Moutakanni, H.~Vo, M.~Szafraniec, V.~Khalidov,
  P.~Fernandez, D.~Haziza, F.~Massa, A.~El-Nouby, et~al.
\newblock Dinov2: Learning robust visual features without supervision.
\newblock \emph{arXiv preprint arXiv:2304.07193}, 2023.

\end{thebibliography}
